# Cross-Modal Domain Adaptation in Brain Disease Diagnosis: Maximum Mean Discrepancy-based Convolutional Neural Networks

Xuran Zhu
School of Information Science & Engineering
Lanzhou University
730000 Lanzhou, China
zhuxr21@lzu.edu.cn

*Abstract*—Brain disorders are a major challenge to global health, causing millions of deaths each year. Accurate diagnosis of these diseases relies heavily on advanced medical imaging techniques such as Magnetic Resonance Imaging (MRI) and Computed Tomography (CT). However, the scarcity of annotated data poses a significant challenge in deploying machine learning models for medical diagnosis. To address this limitation, deep learning techniques have shown considerable promise. Domain adaptation techniques enhance a model's ability to generalize across imaging modalities by transferring knowledge from one domain (e.g., CT images) to another (e.g., MRI images). Such cross-modality adaptation is essential to improve the ability of models to consistently generalize across different imaging modalities. This study collected relevant resources from the Kaggle website and employed the Maximum Mean Difference (MMD) method - a popular domain adaptation method - to reduce the differences between imaging domains. By combining MMD with Convolutional Neural Networks (CNNs), the accuracy and utility of the model is obviously enhanced. The excellent experimental results highlight the great potential of data-driven domain adaptation techniques to improve diagnostic accuracy and efficiency, especially in resource-limited environments. By bridging the gap between different imaging modalities, the study aims to provide clinicians with more reliable diagnostic tools.

## I. INTRODUCTION

Brain diseases are often considered as one of the most dangerous diseases [1]. According to the World Health Statistics 2020 published by WHO, since 2016, more than 10 million people have died from brain diseases every year [2]. Brain diseases are one of the biggest threats to human health. Common brain disorders usually include brain tumors, brain hemorrhages, etc., which are formed for different reasons and in different locations. In this case, when the brain is examined, it is necessary to know if and where it occurs. To solve the problem, many studies introduced Artificial Intelligence (AI) technologies for lesion segmentation, detection-based diagnosis, grading or subtype classification, and outcome prediction [2], which has greatly improved diagnostic efficiency.

And in the past few years, AI technology has realized significant progress and become a hotspot for research and application in many fields [3-6]. Especially in the fields of machine learning and deep learning, a variety of innovative algorithms and models have emerged [7-9]. Machine learning algorithms such as Decision Trees, Random Forests and Support Vector Machines, as well as Convolutional Neural Networks (CNNs) and Artificial Neural Networks in Deep Learning, have demonstrated their powerful capabilities in continuous evolution. These AI models have been used in a wide range of industries, including finance, business and especially medicine [10-12]. In the medical field, the application of AI not only improves the diagnosis and treatment methods of diseases [13, 14], but also plays an important role in disease discovery and other aspects.

One of the important areas of current research is the utilization of artificial intelligence techniques in medical imaging to observe head disorders. Due to the complex and misleading structure of the brain, medical imaging techniques such as Magnetic Resonance Imaging (MRI) and Computed Tomography (CT) are commonly used by physicians to examine brain conditions. These medical imaging modalities can be used to illustrate the internal structure of the brain and describe a three-dimensional model for clinical and research purposes [15].

However, its main challenge in medical image analysis is the lack of labeling data to construct reliable and robust models [16]. An intuitive solution at the moment is to reuse pre-trained models for a number of related domains [17], which is gradually attracting more and more attention. Cheplygina et al [18] provide an extensive review involving semi-supervised learning, multi-instance learning, and transfer learning for medical image analysis. Due to the broad scope of the study, they mainly reviewed transfer learning methods in general and did not specifically focus on domain adaptation. Valverde et al [19] conducted research on the application of transfer learning in magnetic resonance (MR) brain imaging. While Morid et al [20] looked at transfer learning methods based on pre-training on ImageNet that can benefit from knowledge gained from massively labeled natural images. Despite the satisfactory results of their work, they have done relatively little research on model generalizability and there is still a gap in research work on how to transfer models trained with CT images to MRI images.

To address the above issues, this study collected two datasets, brain CT images and brain MRI images, from the Kaggle platform. These datasets mainly include images of diseases such as brain hemorrhage and brain tumor. Through the Maximum Mean Difference (MDD) domain adaptation approach, this study transferred a model from the source domain to the target domain and successfully improved the applicability performance of the model in predicting brain diseases. This approach not only enhances the model's ability to generalize to different medical imaging techniques, but also provides more accurate diagnostic support for patients who are unable to undergo specific types of imaging.

## II. METHODS

### A. Data Preparation

The dataset used in this study was obtained from the Kaggle website and contains a number of MRI and CT images on brain diseases such as brain hemorrhage and brain tumors [21, 22]. These images were divided into two categories, MRI and CT, based on their type, and further differentiated between the presence and absence of disease in each category. Specifically, the number of CT images with disease is 5,841 and the number of CT images without disease is 3,169; the number of MRI images with disease is 1,619 and the number of MRI images without disease is 884. Two sample images are shown in Figure 1.

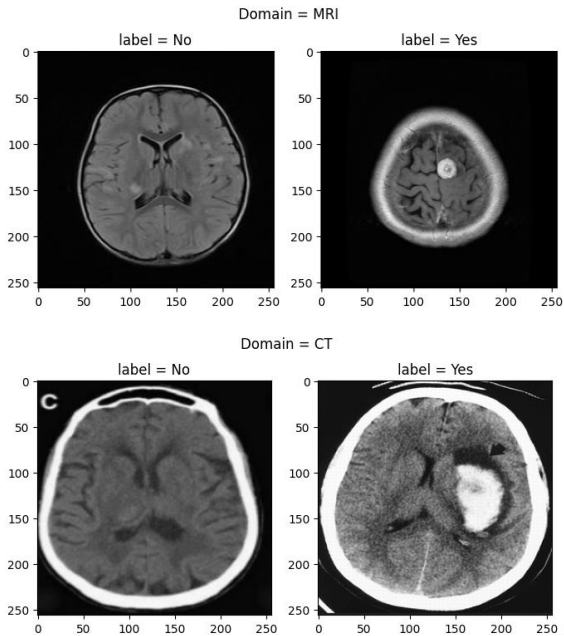

Figure 1. The sample images of MRI and CT collected in this study.

In terms of data processing, this study first converts all the images to RGB three-channel format to ensure that even if the original images are grayscale, they can be unified into a color image format to meet the input requirements of the deep learning model. Also, due to the inconsistent size of the original images, all images were resized to a uniform size of 224×224 pixels. This consistent input shape is crucial to the neural network architecture, allowing for batch processing and reduced computational complexity during training. Finally, to normalize the pixel values, this paper divided the pixel values by 255 to normalize each grayscale image to a range between 0 and 1. This step ensures that the input data to the model has a consistent scale, which helps stabilize training and convergence.

### B. Convolutional Neural Networks for Head Disease Prediction Based on Domain Adaptation

*1) Convolutional Neural Networks:* Convolutional Neural Network is a very effective image processing method commonly used to extract features from images. In this study, the structure of the CNN model consists of several layers. The input layer accepts an RGB image of size 224x224. Meanwhile, the model contains two convolutional layers, the first one uses 16 3x3 filters and the second one uses 32 3x3 filters. Both convolutional layers use the Rectified Linear Unit (ReLU) activation function to ensure nonlinear feature extraction. Each convolutional layer is followed by a 2x2 Maximum Pooling layer to reduce the spatial dimensionality of the features. The Flatten layer then spreads the multidimensional output of the convolutional layers into one dimension for subsequent processing. In the Fully Connected layer, the study added 16 neurons and continued to use the ReLU activation function. To prevent model overfitting, a Dropout layer was set up with a dropout rate of 0.5 to randomly disconnect some of the connections during training. This structure is designed to reduce the risk of overfitting while maintaining the complexity of the model.

*2) Combination of MDD and CNN:* The project employs a fusion of Convolutional Neural Networks (CNNs) and Models of Differential Discrepancy at the Edge (MDD) to address the domain adaptation challenge. First, a CNN model is defined as a feature encoder. This encoder extracts features from the input image through multiple convolutional and pooling layers, and ultimately obtains a compressed feature representation through a fully connected layer. Specifically, the CNN first captures spatial features using convolutional layers, then reduces the spatial dimensionality through pooling layers to improve computational efficiency and reduce the risk of overfitting, and finally reduces the feature dimensionality and enhances the model's generalization ability using fully connected and Dropout layers. Next, a task network is defined to perform the classification task using the features extracted by the encoder. This network contains multiple fully connected layers and utilizes Dropout and maximum paradigm constraints to prevent overfitting and control weight growth. This setup helps to maintain the stability of the network and improve the performance of the model under different data distributions. The MDD model is then introduced as a domain adaptation method. The core idea of MDD is to minimize the difference in distribution between the source and target domains in the task-specific space, i.e., by adapting the model so that the difference between the category boundaries of the two domains is minimized. This is achieved by adding an additional regularization term during training that penalizes the source and target domains for inconsistency in prediction. In the implementation, a custom training callback UpdateLambda is used to dynamically adjust the weight of the regularization term during training. This callback ensures that the strength of the regularization can be appropriately adjusted according to the model's performance on the target domain to achieve the goal of good performance on both domains. Finally, the model is trained on the preprocessed source and target domain data. Through iterative training, the model not only learns effective feature representations and classification rules, but also gradually adapts to the data distribution of the target domain. In this way, the model is able to overcome the differences between domains and improve its generalization ability to new environments.

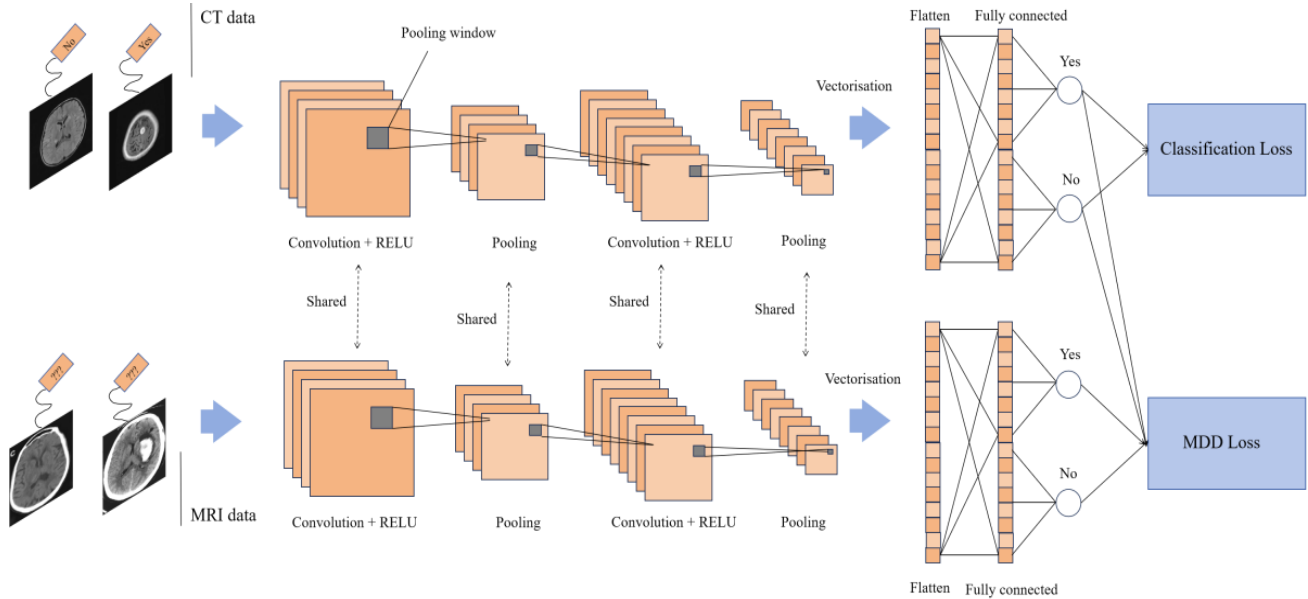

Figure 2. The architecture of the task.

*3) Implementation Details:* During the model training process, this study adopted a series of key parameters and strategies to ensure the stability of the training process and the generalization ability of the model. First, the learning rate of all optimizers was set to 0.0005, this lower learning rate can help the model to adjust parameters stably and avoid excessive oscillation. Meanwhile, the Adam optimizer was selected, which is an optimization algorithm that can automatically adjust the learning rate and is well suited for handling large-scale data and parameter optimization.

To ensure the accuracy of the model in multiclassification tasks, this work uses a categorical cross-entropy loss function. This loss function measures the difference between the probability distribution of the model output and the probability distribution of the true labels and is a common choice for multicategorization tasks. Finally, the study sets the batch size to 16, a relatively small batch size that allows for more frequent parameter updates and thus provides greater flexibility in the training process.

## III. RESULTS AND DISCUSSION

In order to explore the training performance of MDD with different CNN models in the dataset, the experiment repeats multiple sets of experiments after modifying the CNN parameters, focusing on analyzing the metrics including training loss, training accuracy, testing loss, and testing accuracy. Table 1 provides the performance of different model structures.

Table 1. The model performance based on various configurations.

|  | Training Loss | Training Accuracy | Testing Loss | Testing Accuracy |
|---|---|---|---|---|
| Fitting | 0.6337 | 0.6483 | 0.6705 | 0.6134 |
| Mdd (1 CNN layer, filter number: 16) | 0.4163 | 0.7547 | 8.4095 | 0.6316 |
| Mdd (2 CNN layers, filter number:16,32) | 0.4945 | 0.6534 | 0.8294 | 0.6617 |
| Mdd (3 CNN layers, filter number:16,32,64) | 0.6485 | 0.6484 | 0.6696 | 0.6134 |
| Mdd (2 CNN layers, filter number: 4,8) | 0.4175 | 0.7687 | 2.7741 | 0.6418 |
| Mdd (2 CNN layers, filter number: 8,16) | 0.4023 | 0.7699 | 6.7106 | 0.6430 |

Based on the experimental results, it can be observed that different CNN architectures and parameters have a significant impact on the performance of the models. First, a single-layer CNN model (with a channel count of 16) exhibited a high training accuracy of 0.7547, but a testing accuracy of only 0.6316. To explore the generalization ability of the model, this study increased the number of CNN layers to two and set the number of channels to 16 and 32, which resulted in a slight improvement in the test accuracy to 0.6617, while the better test loss implied that the model captured the data features more adequately. Further increasing the number of layers to three and maintaining the number of channels at 16, 32 and 64, the model did not achieve the expected level of accuracy on both training and testing. This suggests that simply increasing the number of layers of the CNN model is not necessarily effective in improving performance. Subsequent experiments set the number of CNN model layers to two and adjusted the number of channels to a lower number of 4 and 8, which resulted in an improvement in training accuracy to 0.7687, but a weak improvement in testing accuracy to 0.6418 accompanied by a higher testing loss. Next, this study tested a dual-layer CNN with a configuration of 8 channels in the first layer and 16 channels in the second layer, a configuration that not only had a high training accuracy of 0.7699, but only a testing accuracy of 0.643 and a relatively high testing loss. So when the number of channels is too low (e.g., 4 and 8), although it helps to reduce overfitting, the model's ability to characterize the model is limited, which may be due to insufficient capture of all valid features in the training data.

Despite optimizing the architecture of the CNN model, the experimental results still show a gap between the desired high accuracy. Future work can be carried out in the following areas using some advanced algorithms in other domains [23-26]:

**Algorithm optimization**: continuous optimization of domain-adaptive algorithms, e.g., by improving the existing MDD technique to more accurately measure and bridge the differences between different data distributions.

**Exploration of data representations**: deep dive into data features and key metrics, especially those that represent differences in data distributions, in order to better design model structures and learning strategies.

**Enhancement of generalization ability**: Explore more ways to improve the generalization ability of the model, such as data enhancement and regularization strategies.

IV. CONCLUSION

This paper effectively leverages a fusion of deep learning and transfer learning techniques, particularly utilizing the MDD method, to enhance model performance when confronted with new data. Through a meticulous series of experiments, the study scrutinizes the influence of various CNN architectures and parameters on model efficacy. Notably, it is demonstrated that by judiciously adjusting the CNN model's layer count and channel configurations, substantial enhancements in both training and test accuracy can be achieved.

Despite these advancements, it's acknowledged that the current model's accuracy remains below desired thresholds. Consequently, future investigations will delve into novel modules or methodologies, such as the integration of auxiliary mechanisms or novel evaluation metrics, aimed at further amplifying the model's performance.